\documentclass[10pt,twocolumn,letterpaper]{article}

\usepackage{iccv}
\usepackage{times}
\usepackage{epsfig}
\usepackage{graphicx}
\usepackage{amsmath}
\usepackage{amssymb}
\usepackage{hyperref}
\usepackage{url}
\usepackage{amsmath}
\usepackage{amssymb}
\usepackage{graphicx}
\usepackage{caption}
\usepackage{subcaption}
\usepackage{epstopdf}

 \iccvfinalcopy 

\def\iccvPaperID{1120} 

\ificcvfinal\fi
\begin{document}

\title{Unsupervised Learning of Spatiotemporally Coherent Metrics}

\author{Ross Goroshin\\
Courant Institute, NYU\\
{\tt\small goroshin@cs.nyu.edu}
\and 
Joan Bruna\\ 
University of California, Berkeley\\
{\tt\small joan.bruna@berkeley.edu} 
\and
Jonathan Tompson\\
Google Inc.\\
{\tt\small tompson@google.com}
\and
David Eigen\\
Courant Institute, NYU\\
{\tt\small deigen@cs.nyu.edu}
\and
Yann LeCun\\
Courant Institute, NYU \& Facebook AI Research \\
{\tt\small yann@cs.nyu.edu} 
}

\maketitle

\begin{abstract}
   Current state-of-the-art classification and detection algorithms train deep convolutional networks using labeled data. In this work we study unsupervised feature learning with convolutional networks in the context of temporally coherent unlabeled data. We focus on feature learning from unlabeled video data, using the assumption that adjacent video frames contain semantically similar information. This assumption is exploited to train a convolutional pooling auto-encoder regularized by slowness and sparsity priors. We establish a connection between slow feature learning and metric learning. Using this connection we define "temporal coherence"--a criterion which can be used to set hyper-parameters in a principled and automated manner. In a transfer learning experiment, we show that the resulting encoder can be used to define a more semantically coherent metric without the use of labels.
\end{abstract}

\section{Introduction}
Is it possible to characterize ``good'' representations without specifying a task a priori? If so, does there exist a set of generic priors which lead to such representations? In recent years state-of-the-art results from supervised learning suggest that the most powerful representations for solving specific tasks can be learned from the data itself. It has been hypothesized that large collections of unprocessed and unlabeled data can be used to learn generically useful representations. However the principles which would lead to these representations in the realm of unsupervised learning remain elusive. Temporal coherence is a form of weak supervision, which we exploit to learn generic signal representations that are stable with respect to the variability in natural video, including local deformations. 

Our main assumption is that data samples that are temporal neighbors are also likely to be neighbors in the latent space. For example, adjacent frames in a video sequence are more likely to be semantically similar than non-adjacent frames. This assumption naturally leads to the slowness prior on features which was introduced in SFA (\cite{SFA}). 

This prior has been successfully applied to metric learning, as a regularizer in supervised learning, and in unsupervised learning (\cite{DrLIM,DrLIMVideo,SFA}). A popular assumption in unsupervised learning is that high dimensional data lies on a low dimensional manifold  parametrized by the latent variables as in \cite{Bengio2012,CAE,DAE,SATAE}. In this case, temporal sequences can be thought of as one-dimensional trajectories on this manifold. Thus, an ensemble of sequences that pass through a common data sample have the potential to reveal the local latent variable structure within a neighborhood of that sample. 

Non-linear operators consisting of a redundant linear transformation followed by a point-wise nonlinearity and a local pooling, are fundamental building blocks in deep convolutional networks. This is due to their capacity to generate local invariance while preserving discriminative information (\cite{LeCun1998, JoanScat}). We justify that pooling operators are a natural choice for our unsupervised learning architecture since they induce invariance to local deformations. The resulting pooling auto-encoder model captures the main source of variability in natural video sequences, which can be further exploited by enforcing a convolutional structure. Experiments on YouTube data show that one can learn pooling representations with good discrimination and stability to observed temporal variability. We show that these features can be used to define a semantically coherent metric which we evaluate on temporal and class-based retrieval tasks. 

\section{Contributions and Prior Work} 
The problem of learning temporally stable representations has been extensively studied in the literature, most prominently in Slow Feature Analysis (SFA) and Slow Subspace Analysis (SSA) (\cite{SFA,SSA,hyvarinen2003bubbles}). Works that learn slow features distinguish themselves mainly in three ways: (1) how the features are parametrized, (2) how the trivial (constant) solution is avoided, and (3) whether or not additional priors such as independence or sparsity are imposed on the learned features. 

The features presented in SFA take the form of a nonlinear transformation of the input, specifically a quadratic expansion followed by a linear combination using learned weights optimized for slowness (\cite{SFA}). 
This parametrization is equivalent to projecting onto a learned basis followed by $L_2$ pooling. The recent work by \cite{complexCells} uses features which are composed of projection onto a learned unitary basis followed by a local $L_2$ pooling in groups of two. 

Slow feature learning methods also differ in the way that they avoid the trivial solution of learning to extract constant features. Constant features are perfectly slow (invariant), however they are not informative (discriminative) with respect to the input. All slow feature learning methods must make a trade-off between the discriminability and stability of the learned features in order to avoid trivial solutions. \emph{To the best of our knowledge, our work is the first to suggest how to make this trade-off in a principled manner.} Slow Feature Analysis introduces two additional constraints, namely that the learned features must have unit variance and must be decorrelated from one another. In the work by \cite{complexCells}, the linear part of the transformation into feature space is constrained to be unitary. Enforcing that the transform be unitary implies that it is invertible \emph{for all inputs}, and not just the data samples. This unnecessarily limits the invariance properties of the transform and precludes the possibility of learning over-complete bases. Since the pooling operation following this linear transform has no trainable parameters, including this constraint is sufficient to avoid the trivial solution. Metric learning approaches (\cite{DrLIM}) can be used to perform dimensionality reduction by optimizing a criteria which minimizes the distance between temporally adjacent samples in the transformed space, while repelling non-adjacent samples with a hinge loss, as explained in Section \ref{sec:slowmetric}. The margin based contrastive term in DrLIM is explicitly designed to only avoid the constant solution and provides no guarantee on how informative the learned features are. Furthermore since distances grow exponentially due to the curse of dimensionality, metric based contrastive terms can be trivially satisfied in high dimensions.

Our approach uses a reconstruction criterion as a contrastive term. This approach is most similar to the one taken by \cite{groupSparsity} when optimizing group sparsity. In this work group-sparsity is replaced by slowness, and multiple layers of convolutional slow features are trained. 

Several other studies combine the slowness prior with independence inducing priors \cite{complexCells, Cadieu, zou2012deep}. For a detailed discussion on the connection between independence and sparsity see \cite{Huyvarinen}. However, our model maximizes the sparsity of the representation \emph{before} the pooling operator. This allows our model to be interpreted as a sparse auto-encoder additionally regularized by slowness through a local pooling operator.   

In this work we introduce the use of convolutional pooling architectures for slow feature learning.
At small spatial scales, local translations comprise the dominant source of variability in natural video; this is why many previous works on slowness learn mainly locally translation-invariant features (\cite{SFA,SSA,complexCells}). However, convolutional pooling architectures are locally translation-invariant by design, which allows our model to learn features that capture a richer class of invariances, beyond translation. Finally, we demonstrate that nontrivial convolutional dictionaries can be learned in the unsupervised setting using only stochastic gradient descent (on mini-batches), despite their huge redundancy --- that is, without resorting to alternating descent methods or iterative sparse inference algorithms. 

\section{Slowness as Metric Learning }
\label{sec:slowmetric}
\begin{figure}
\centering
\begin{subfigure}[b]{0.49 \textwidth}
\centering
\includegraphics[scale=0.5]{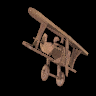} 
\includegraphics[scale=0.5]{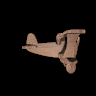} 
\includegraphics[scale=0.5]{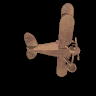}
        \caption{}
\label{fig:toyplane}
\end{subfigure}
%
\begin{subfigure}[b]{0.49 \textwidth}
\centering
\includegraphics[scale=0.5,trim = 120 100 130 70, clip]{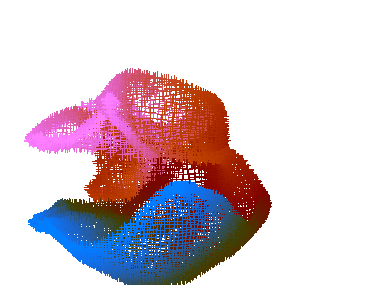}
\includegraphics[scale=0.5,trim = 120 100 130 70, clip]{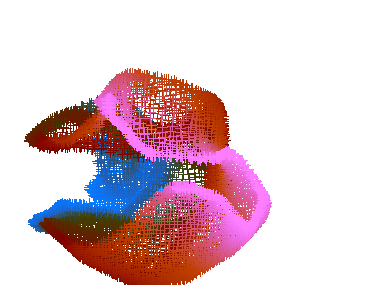}\\
\includegraphics[scale=0.5,trim = 120 100 130 70, clip]{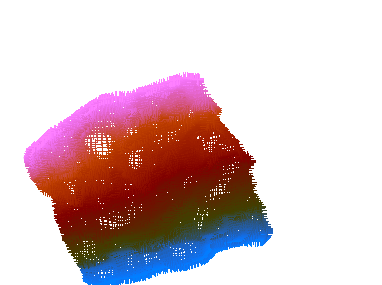} 
\includegraphics[scale=0.5,trim = 120 100 130 70, clip]{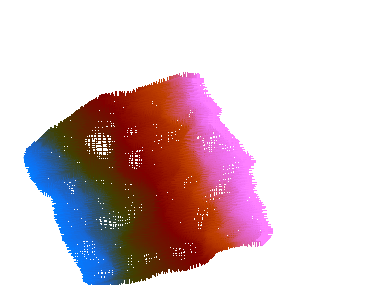}
        \caption{}
        \label{fig:drlim} 
\end{subfigure}

\caption{(a) Three samples from our rotating plane toy dataset. (b) Scatter plot of the dataset plotted in the output space of $G_W$ at the start (top) and end (bottom) of training. The left side of the figure is colored by the yaw angle, and the right side by roll, $0^{\circ}$  blue, $90^{\circ}$ in pink.}  
\end{figure} 
Temporal coherence can be exploited by assuming a prior on the features extracted from the temporal data sequence. One such prior is that the features should vary slowly with respect to time. In the discrete time setting this prior corresponds to minimizing an $L^p$ norm of the difference of feature vectors for temporally adjacent inputs. 
Consider a video sequence with $T$ frames, if $z_t$ represents the feature vector extracted from the frame at time $t$ then the slowness prior corresponds to minimizing $\sum_{t=1}^{T} \| z_t - z_{t-1} \|_p$. To avoid the degenerate solution $z_t = z_0 ~\mbox{for}~ t=1...T$, a second term is introduced which encourages data samples that are \emph{not} temporal neighbors to be separated by at least a distance of $m$-units in feature space, where $m$ is known as the margin. In the temporal setting this corresponds to minimizing $max(0,m-\|z_t - z_{t'}\|_p)$, where $|t-t'| > 1$. Together the two terms form the loss function introduced in \cite{DrLIM} as a dimension reduction and data visualization algorithm known as DrLIM. 
Assume that there is a differentiable mapping from input space to feature space which operates on \emph{individual} temporal samples. Denote this mapping by $G$ and assume it is parametrized by a set of trainable coefficients denoted by $W$. That is, $z_t = G_W(x_t)$. The per-sample loss function can be written as: 
\begin{equation} 
\label{eqn:drlimcrit}
\begin{split} 
&L(x_t,x_{t'},W)= \\ 
			 &\left\{
                \begin{array}{ll}
                 \|G_W(x_t) - G_W(x_{t'})\|_p, &\text{if}~|t-t'| = 1  \\
                 \max(0,m-\|G_W(x_t) - G_W(x_{t'})\|_p) &\text{if}~|t-t'| > 1
                \end{array}
              \right.
\end{split} 
\end{equation} 
In practice the above loss is minimized by constructing a "Siamese" network (\cite{siamese}) with shared weights whose inputs are pairs of samples along with their temporal indices. The loss is minimized with respect to the trainable parameters using stochastic gradient descent via back-propagation. To demonstrate the effect of minimizing Equation \ref{eqn:drlimcrit} on temporally coherent data, consider a toy data-set consisting of only one object. The data-set is generated by rotating a 3D model of a toy plane (Figure \ref{fig:toyplane}) by $90^{\circ}$ in one-degree increments around two-axes of rotation, generating a total of 8100 data samples. Input images ($96 \times 96$) are projected into two-dimensional output space by the mapping $G_W$. In this example the mapping $G_W(X): \mathbb{R}^{9216} \to \mathbb{R}^2 $. We chose $G_W$ to be a fully connected two layer neural network. In effect this data-set lies on an intrinsically two-dimensional manifold parametrized by two rotation angles. Since the sequence was generated by continuously rotating the object, temporal neighbors correspond to images of the object in similar configurations. 
Figure \ref{fig:drlim} shows the data-set plotted in the output space of $G_W$ at the start (top row) and end (bottom row) of training. The left and right hand sides of Figure~\ref{fig:drlim} are colorized by the two rotational angles, which are never explicitly presented to the network. This result implies that $G_W$ has learned a mapping in which the latent variables (rotation angles) are linearized. Furthermore, the gradients corresponding to the two rotation angles are nearly orthogonal in the output space, which implies that the two features extracted by $G_W$ are independent. 
\section{Slow Feature Pooling Auto-Encoders}
\label{sfautoencs}
The second contrastive term in Equation \ref{eqn:drlimcrit} only acts to avoid the degenerate solution in which $G_W$ is a constant mapping, it does not guarantee that the resulting feature space is informative with respect to the input. 
This discriminative criteria only depends on pairwise distances in the representation space which is a geometrically weak notion in high
dimensions. We propose to replace this contrastive term with a term that penalizes the reconstruction error of both data samples. 
Introducing a reconstruction terms not only prevents the constant solution but also acts to explicitly preserve information about the input. 
This is a useful property of features which are obtained using unsupervised learning; since the task to which these features will be applied 
is not known a priori, we would like to preserve as much information about the input as possible. 

\begin{figure*}[twocolumn]
\centering
\includegraphics[scale=0.75,trim = 15 350 290 39, clip]{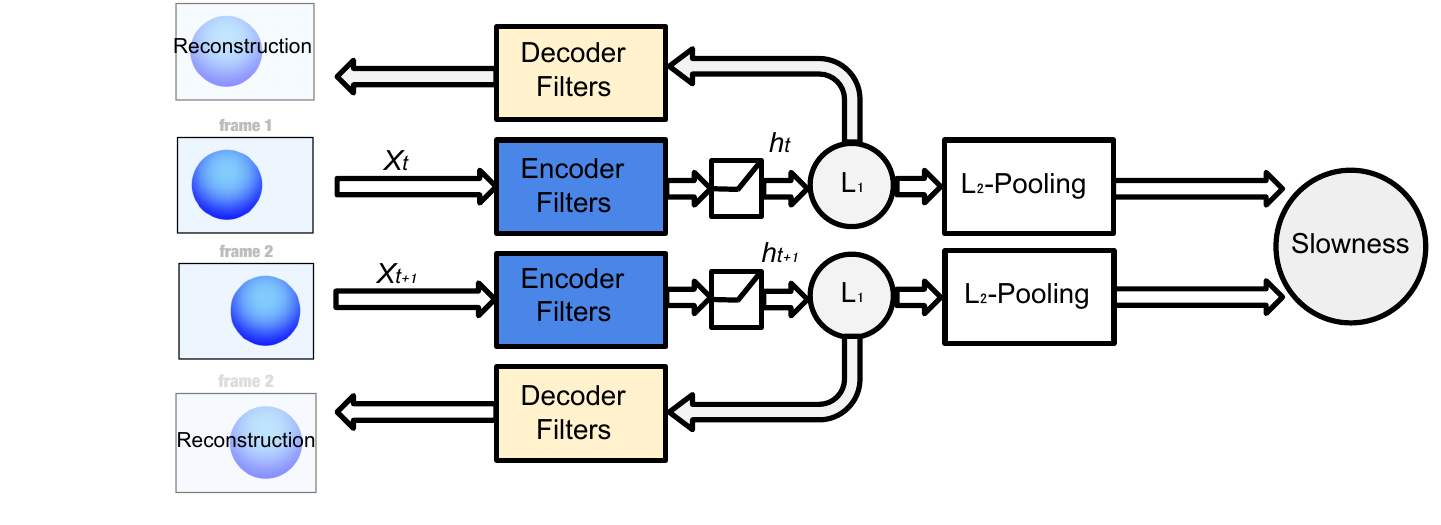}
\caption{Block diagram of the Siamese convolutional model trained on pairs of frames. \label{fig:diagram}}  
\end{figure*}

What is the optimal architecture of $G_{W}$ for extracting slow features? Slow features are invariant to temporal changes by definition. In natural video and on small spatial scales these changes mainly correspond to local translations and deformations. Invariances to such changes can be achieved using appropriate pooling operators \cite{JoanScat,LeCun1998}. 
Such operators are at the heart of deep convolutional networks (ConvNets), currently the most successful supervised feature learning architectures \cite{ImageNet}. Inspired by these observations, let $G_{W_e}$ be a two stage encoder comprised of a learned, generally over-complete, linear map ($W_e$) and rectifying nonlinearity $f(\cdot)$, followed by a local pooling. Let the $N$ hidden activations, $h = f(W_ex)$, be subdivided into $K$ potentially overlapping neighborhoods denoted by $P_i$. Note that biases are absorbed by expressing the input $x$ in homogeneous coordinates. Feature $z_i$ produced by the encoder for the input at time $t$ can be expressed as $G_{W_e}^i(t) = \|h_t\|^{P_i}_p =\left(\sum_{j \in P_i} h_{tj}^{p} \right)^{\frac{1}{p}}$. Training through a local pooling operator enforces a local topology on the hidden activations, inducing units that are pooled together to learn complimentary features. In the following experiments we will use $p=2$. Although it has recently been shown that it is possible to recover the input when $W_e$ is sufficiently redundant, reconstructing from these coefficients corresponds to solving a phase recovery problem \cite{JoanPooling} which is not possible with a simple inverse mapping, such as a linear map $W_d$. Instead of reconstructing from $z$ we reconstruct from the hidden representation $h$. This is the same approach taken when training group-sparse auto-encoders \cite{groupSparsity}. In order to promote sparse activations in the case of over-complete bases we additionally include a sparsifying $L_1$ penalty on the hidden activations. Including the rectifying nonlinearity becomes critical for learning sparse inference in a hugely redundant dictionary, e.g. convolutional dictionaries \cite{LISTA}. The complete loss function is: 
\begin{equation}
\label{eqn:loss}
\begin{split} 
L(x_t,x_{t'},W)= \sum_{\tau = \{t,t'\}} \left(\|W_d h_\tau - x_\tau\|^2 + \alpha|h_\tau| \right) \\ +~\beta \sum_{i=1}^K \left| \|h_t \|^{P_i} - \|h_{t'}\|^{P_i} \right|
\end{split} 
\end{equation} 
Figure \ref{fig:diagram} shows a convolutional version of the proposed architecture and loss. 
This combination of loss and architecture can be interpreted as follows: the sparsity penalty induces the first stage of the encoder, $h=f(W_ex)$, to approximately infer sparse codes in the analysis dictionary $W_e$; the slowness penalty induces the formation of pool groups whose output is stable with respect to temporal deformations. In other words, the first stage partitions the input space into potentially disjoint linear subspaces and the second stage (pooling) recombines these partitions into temporally stable groups. This can be seen as a sparse auto-encoder whose pooled codes are additionally regularized by slowness.

\begin{figure}
  \centering
  \begin{subfigure}[b]{0.45\textwidth}
        \includegraphics[width=\textwidth]{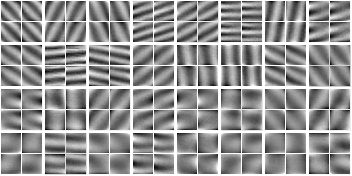}
        \caption{}
        \label{fig:pooldec}
  \end{subfigure}
  \begin{subfigure}[b]{0.45\textwidth}
        \includegraphics[width=\textwidth]{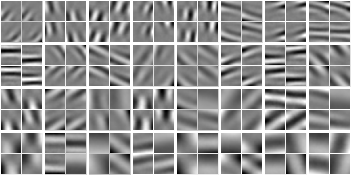}
        \caption{}
        \label{fig:pooll1dec}
  \end{subfigure}
  \caption{Pooled decoder dictionaries learned without (a) and with (b) the $L_1$ penalty using (\ref{eqn:loss}).}
  \label{fig:sfpool}
\end{figure}

\subsection{Fully-Connected Architecture}
To gain an intuition for the properties of the minima of Equation \ref{eqn:loss} for natural data, an auto-encoder was trained on a small dataset consisting of natural movie patches. This data set consists of approximately 170,000, $20 \times 20$ gray scale patches extracted from full resolution movies. 
Minimizing Equation \ref{eqn:loss} with $\alpha=0$ results in the learned decoder basis shown in Figure \ref{fig:pooldec}. Here a dictionary of 512 basis elements was trained whose outputs were pooled in non-overlapping groups of four resulting in 128 output features. Only the slowest 32 groups are shown in Figure \ref{fig:pooldec}. The learned dictionary has a strong resemblance to the two-dimensional Fourier basis, where most groups are comprised of phase shifted versions of the same spatial frequency. Since translations are an invariant of the local modulus of the Fourier transform, the result of this experiment is indicative of the fact that translations are the principal source of variation at small spatial scales. Minimizing Equation \ref{eqn:loss} with $\alpha > 0$ results in a more localized basis depicted in Figure \ref{fig:pooll1dec}. This basis is more consistent with a local deformation model as opposed to a global one. 
\subsection{Convolutional Architecture}
By replacing all linear operators in our model with convolutional filter banks and including spatial pooling, translation invariance need not be learned \cite{LeCun1998}. In all other respects the convolutional model is conceptually identical to the fully connected model described in the previous section. One important difference between fully-connected and convolutional dictionaries is that the latter can be massively over-complete, making sparse inference potentially more challenging. Nevertheless we found that non-trivial dictionaries (see Figure \ref{fig:slowfilters}) can be learned using purely stochastic optimization, that is, without a separate sparse inference phase. Let the linear stage of the encoder consist of a filter bank which takes $C$ input feature maps (corresponding to the 3 color channels for the first stage) and produces $D$ output feature maps. Correspondingly, the convolutional decoder transforms these $D$ feature maps back to $C$ color channels. In the convolutional setting slowness is measured by subtracting corresponding spatial locations in temporally adjacent feature maps. In order to produce slow features a convolutional network must compensate for the motion in the video sequence by producing \emph{spatially} aligned activations for \emph{temporally} adjacent samples. In other words, in order to produce slow features the network must implicitly learn to track common patterns by learning features which are invariant to the deformations exhibited by these patterns in the temporal sequence. The primary mechanism for producing these invariances is pooling in space and across features \cite{MaxOut}. Spatial pooling induces local translation invariance. Pooling across feature maps allows the network to potentially learn feature groups that are stable with respect to more general deformations. Intuitively, maximizing slowness in a convolutional architecture leads to \emph{spatiotemporally} coherent features. 
\section{Experimental Results}
To verify the connection between slowness and metric learning, we evaluate the metric properties of the learned features. It is well known that distance in the extrinsic (input pixel) space is not a reliable measure of semantic similarity. Maximizing slowness corresponds to minimizing the distance between adjacent frames in code space, therefore neighbors in code space should correspond to temporal neighbors. This claim can be tested by computing the nearest neighbors to a query frame in code space, and verifying whether they correspond to its temporal neighbors. However, the features must also be discriminative so as not to collapse temporally distant samples. In order to make this trade-off in a principled manner, a dataset comprised of short natural scenes was collected. Hyper-parameters are selected which maximize the so called "temporal coherence" of the features which define the metric. We define the temporal coherence of a metric $G_W(\cdot)$ as the area under the precision-recall curve, where precision is defined as the proportion of the nearest neighbors that come from the same scene, and recall is defined as the proportion of frames recalled from that scene. In our experiments, we used the middle frame from each scene as the query. 

\begin{figure}
\centering
\includegraphics[scale=0.5]{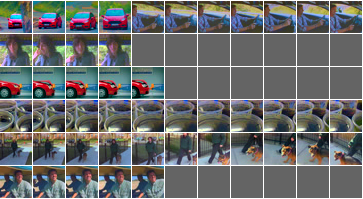}
\caption{Six scenes from our YouTube dataset \label{fig:youtube}}  
\end{figure}

However, temporal coherence can be a very weak measure of discriminability; it merely requires that scenes be easy to disambiguate in feature space. If the scenes are quite distinct, then maximizing temporal coherence directly can lead to weakly discriminative features (e.g.~color histograms can exhibit good temporal coherence). We therefore evaluate the learned features on a more demanding task by assessing how well the metric learned from the YouTube dataset transfers to a classification task on the CIFAR-10 dataset. Average class-based precision is measured in feature space by using the test set as the query images and finding nearest neighbors in the training set. Precision is defined as the proportion of nearest neighbors that have the same label. As on the YouTube dataset we evaluate the average precision for the nearest 40 neighbors. The CIFAR dataset contains considerably more interclass variability than the scenes in our YouTube dataset, nevertheless many class instances are visually similar.  
Our training dataset consists of approximately $150,000$ frames extracted from YouTube videos. Of these, approximately $20,000$ frames were held out for testing. The training and test set frames were collected from separate videos. The videos were automatically segmented into scenes of variable length (2-40 frames) by detecting large $L_2$ changes between adjacent frames. Each color frame was down-sampled to a $32 \times 32$ spatial resolution and the entire dataset was ZCA whitened \cite{alexthesis}. Six scenes from the test set are shown in Figure \ref{fig:youtube} where the first scene (top row) is incorrectly segmented.   

We compare the features learned by minimizing the loss in Equation \ref{eqn:loss} with the features learned by minimizing DrLIM (Equation \ref{eqn:drlimcrit}) and group sparsity (Equation \ref{eqn:groupL1}) losses. Once trained, the convolution, rectification, and pooling stages are used to transform the dataset into the feature space. We use cosine distance in feature space to determine the nearest neighbors and select hyperparmeters for each method which maximize the temporal coherence measure.

We trained two layers of our model using greedy layer-wise training \cite{Bengio2012}. The first layer model contains a filter bank consisting of 64 kernels with $9 \times 9$ spatial support. The first $L_2$ pooling layer computes the local modulus volumetrically, that is \emph{across} feature maps in non-overlapping groups of four and spatially in $2 \times 2$ non-overlapping neighborhoods. Thus the output feature vector of the first stage ($z_1$) has dimensions $16 \times 16 \times 16$ (4096). Our second stage consists of 64 $5 \times 5$ convolutional filters, and performs $4 \times 4$ spatial pooling producing a second layer code ($z_2$) of dimension $64 \times 4 \times 4$ (1024). The output of the second stage corresponds to a dimension reduction by a factor of three relative to the input dimension.

Identical one and two-layer architectures were trained using the group sparsity prior, similar to \cite{groupSparsity}. As in the slowness model, the two layer architecture was trained greedily. Using the same notation as Equation \ref{eqn:loss}, the corresponding loss can be written as:  
\begin{equation}
L(x_t,W)= \sum_{\tau} \|W_d h_\tau - x_\tau\|^2 +\alpha \|h_\tau \|^{P_i} 
\label{eqn:groupL1}
\end{equation} 
Finally, identical one and two-layer architectures were also trained by minimizing the DrLIM loss in Equation \ref{eqn:drlimcrit}. Negative pairs, corresponding to temporally non-adjacent frames, were independently selected at random. In order to achieve the best temporal precision-recall performance, we found that each mini-batch should consist of a large proportion of negative to positive samples (at least five-to-one). Unlike the auto-encoder methods, the two layer architecture was trained jointly rather than greedily.
\begin{figure*}[ht]
  \centering
  \newcommand{\figspace}{0.0cm}
  \newcommand{\figsize}{0.24\textwidth}
   \begin{subfigure}[b]{\figsize}
        \includegraphics[width=\textwidth]{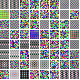}
        \caption{}
        \label{fig:drlimfilters}
  \end{subfigure}
   \hspace{\figspace} 
  \begin{subfigure}[b]{\figsize}
		\includegraphics[width=\textwidth]{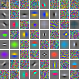}
        \caption{}
        \label{fig:L1filters}
  \end{subfigure}
  \hspace{\figspace} 
  \begin{subfigure}[b]{\figsize}
  		\includegraphics[width=\textwidth]{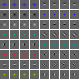}
        \caption{}
        \label{fig:groupL1filters}
  \end{subfigure}
   \hspace{\figspace} 
  \begin{subfigure}[b]{\figsize}
        \includegraphics[width=\textwidth]{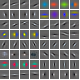}
        \caption{}
        \label{fig:slowfilters}
  \end{subfigure}
	\caption{Pooled convolutional dictionaries (decoders) learned with: (a) DrLIM and (b) sparsity only, (c) group sparsity, and (d) sparsity and slowness. Groups of four features that were pooled together are depicted as horizontally adjacent filters.}
  \label{fig:filters}
\end{figure*}

\begin{figure}
  \center
  \begin{subfigure}[b]{0.45\textwidth}
  \includegraphics[width=\textwidth, trim = 0 0 34 0, clip]{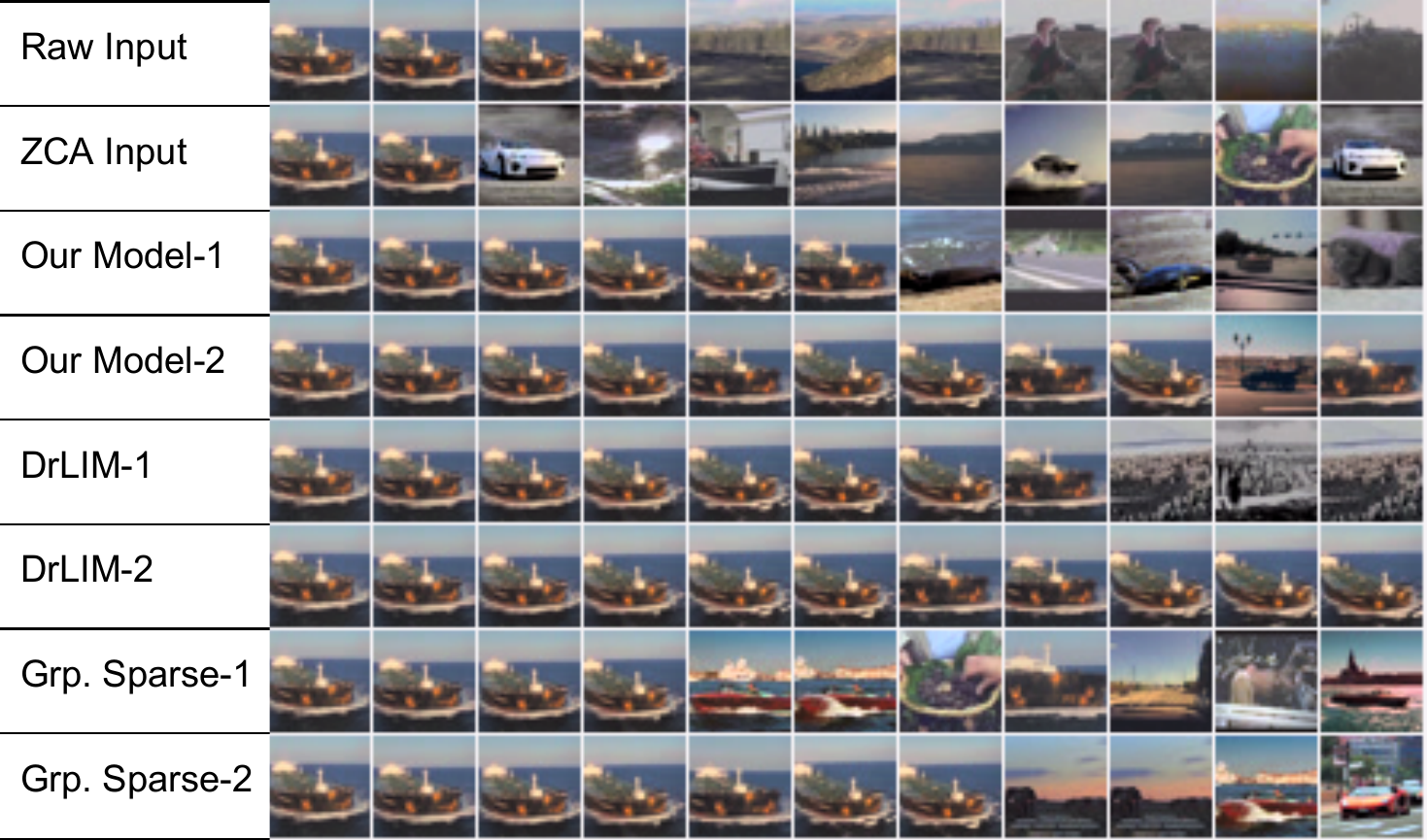}
   \includegraphics[width=\textwidth, trim = 0 0 34 0, clip]{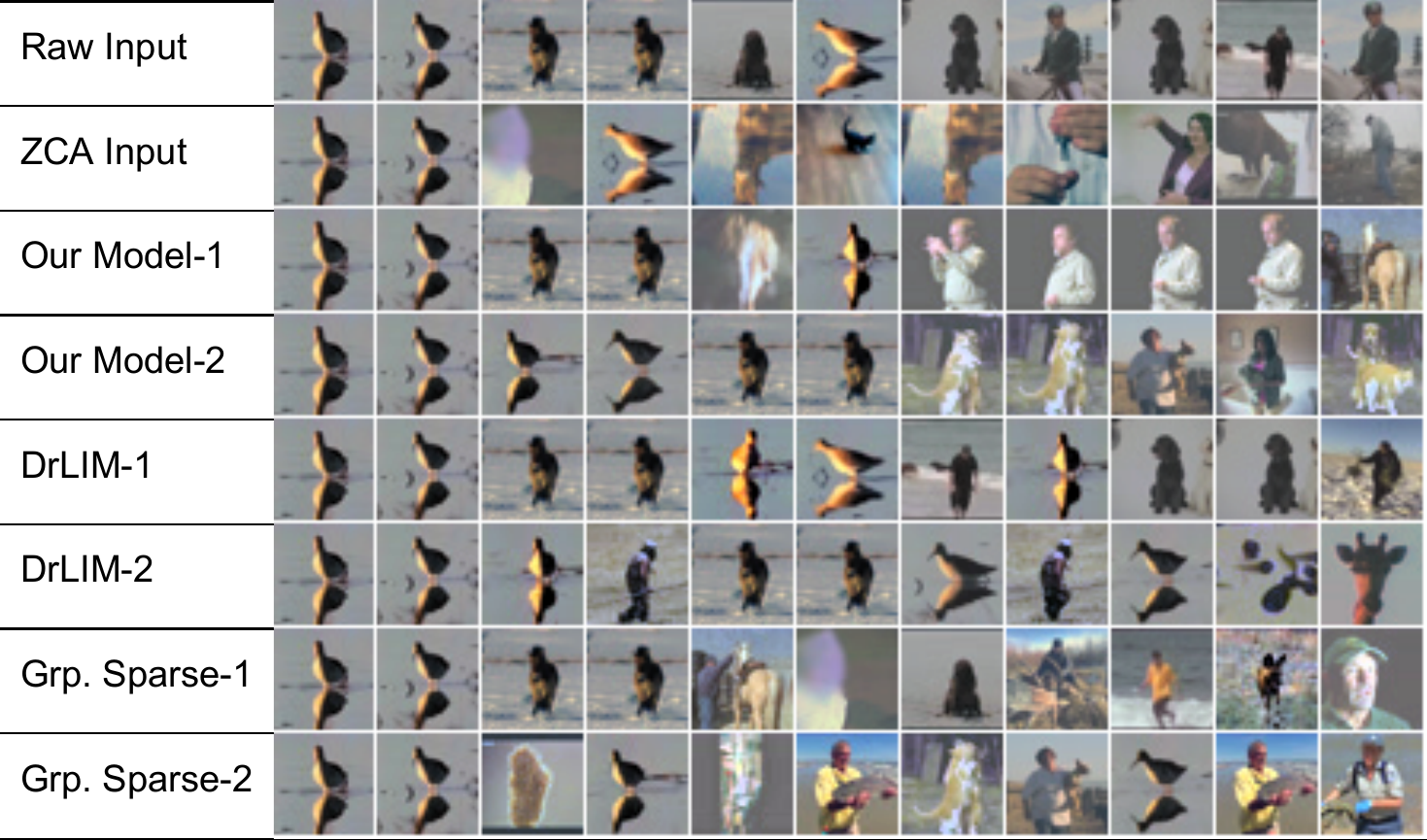}
        \caption{}
        \label{fig:videoquery}
  \end{subfigure}
  \centering \hspace{0.1cm}
  \begin{subfigure}[b]{0.45\textwidth}
       \includegraphics[width=\textwidth, trim = 0 0 34 0, clip]{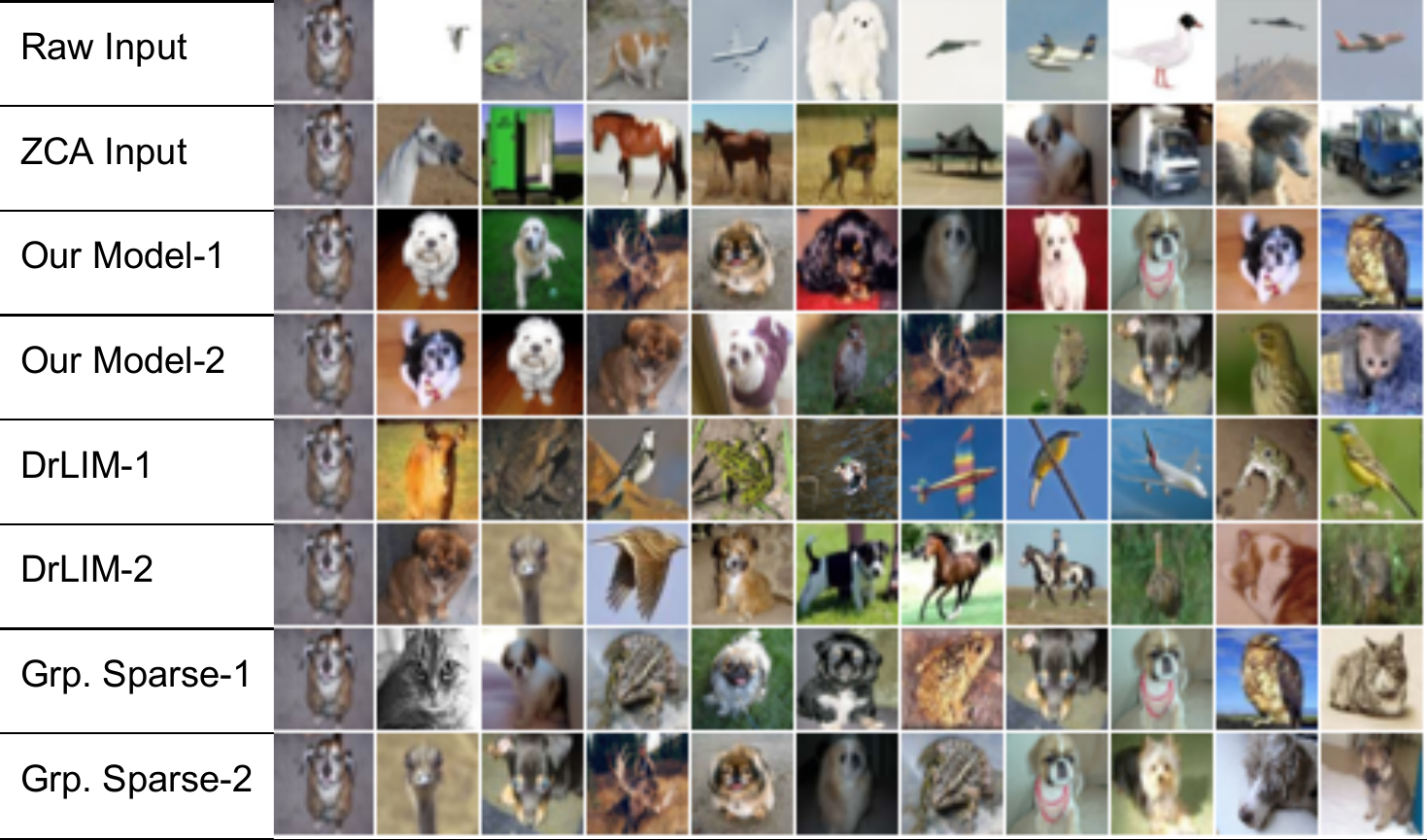}
       \includegraphics[width=\textwidth, trim = 0 0 34 0, clip]{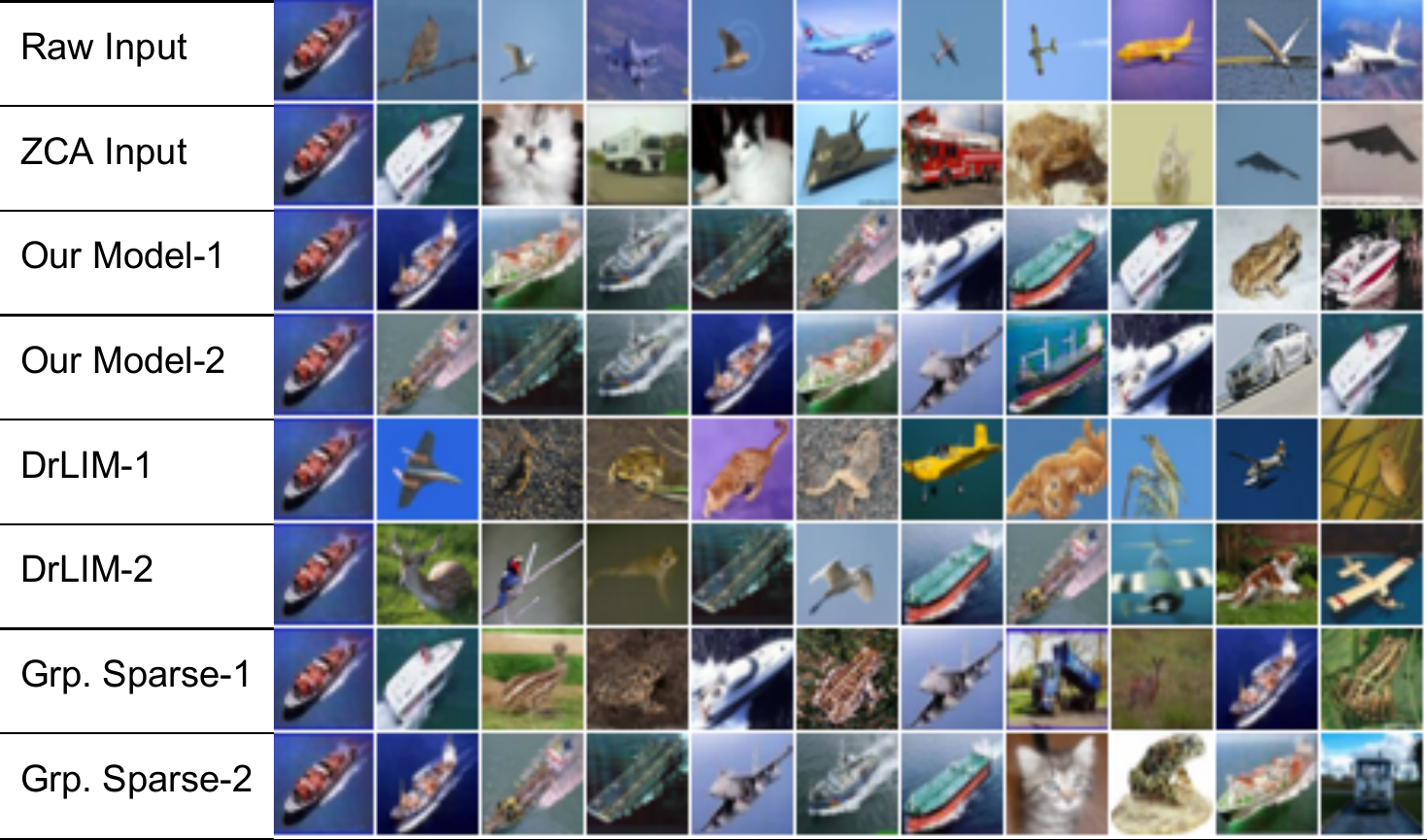}
        \caption{}
        \label{fig:cifarquery}
  \end{subfigure}
  \caption{Query results in the (a) video and (b) CIFAR-10 datasets. Each row corresponds to a different feature space in which the queries were performed; numbers (1 or 2) denote the number of convolution-pooling layers. \label{fig:query}}  
   \label{fig:query}
\end{figure}

\begin{figure}
  \centering
  \begin{subfigure}[b]{0.45\textwidth}
        \includegraphics[width=\textwidth]{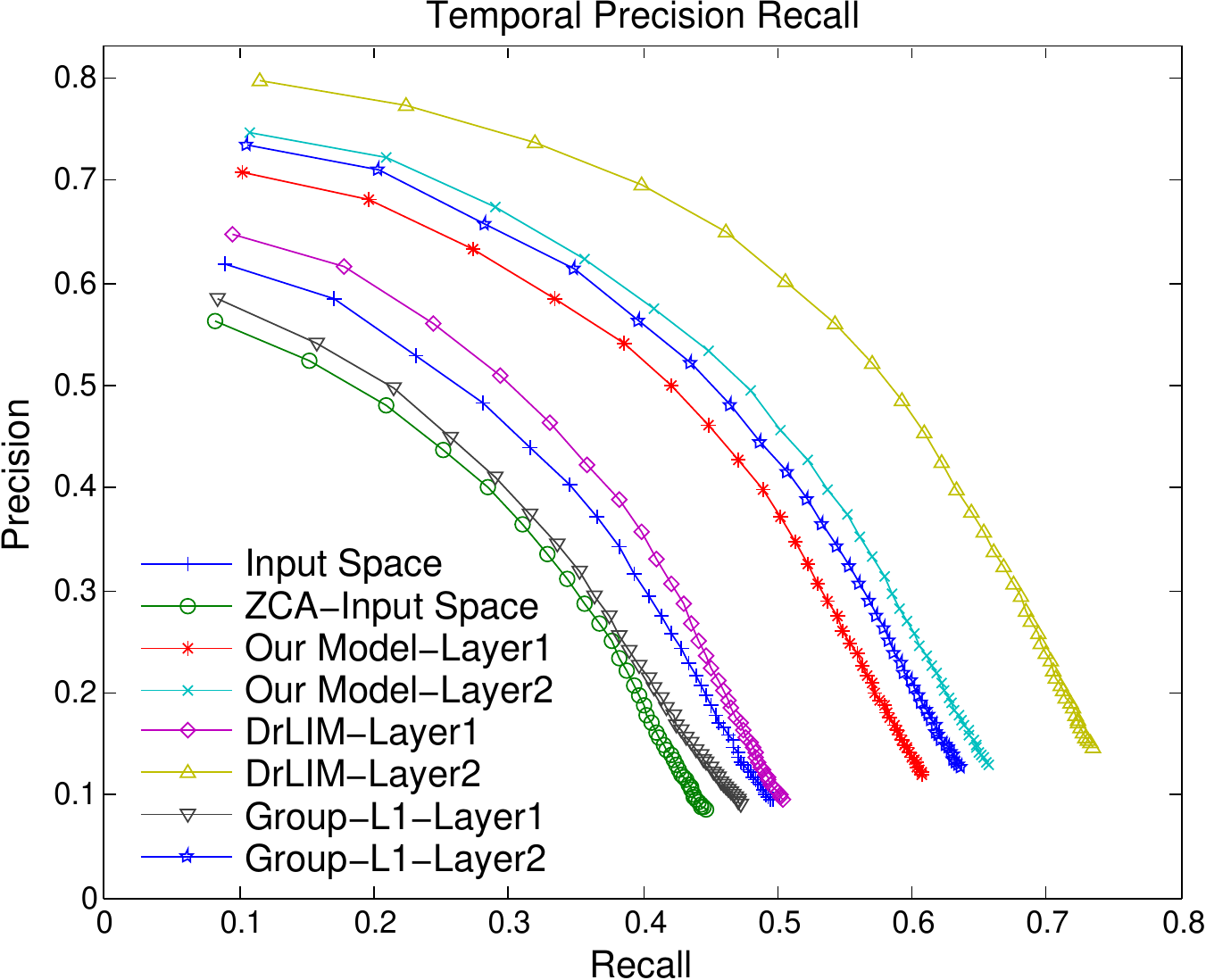}
        \caption{}
        \label{fig:ROCtime}
  \end{subfigure} 
  \centering
  \begin{subfigure}[b]{0.45\textwidth}
        \includegraphics[width=\textwidth]{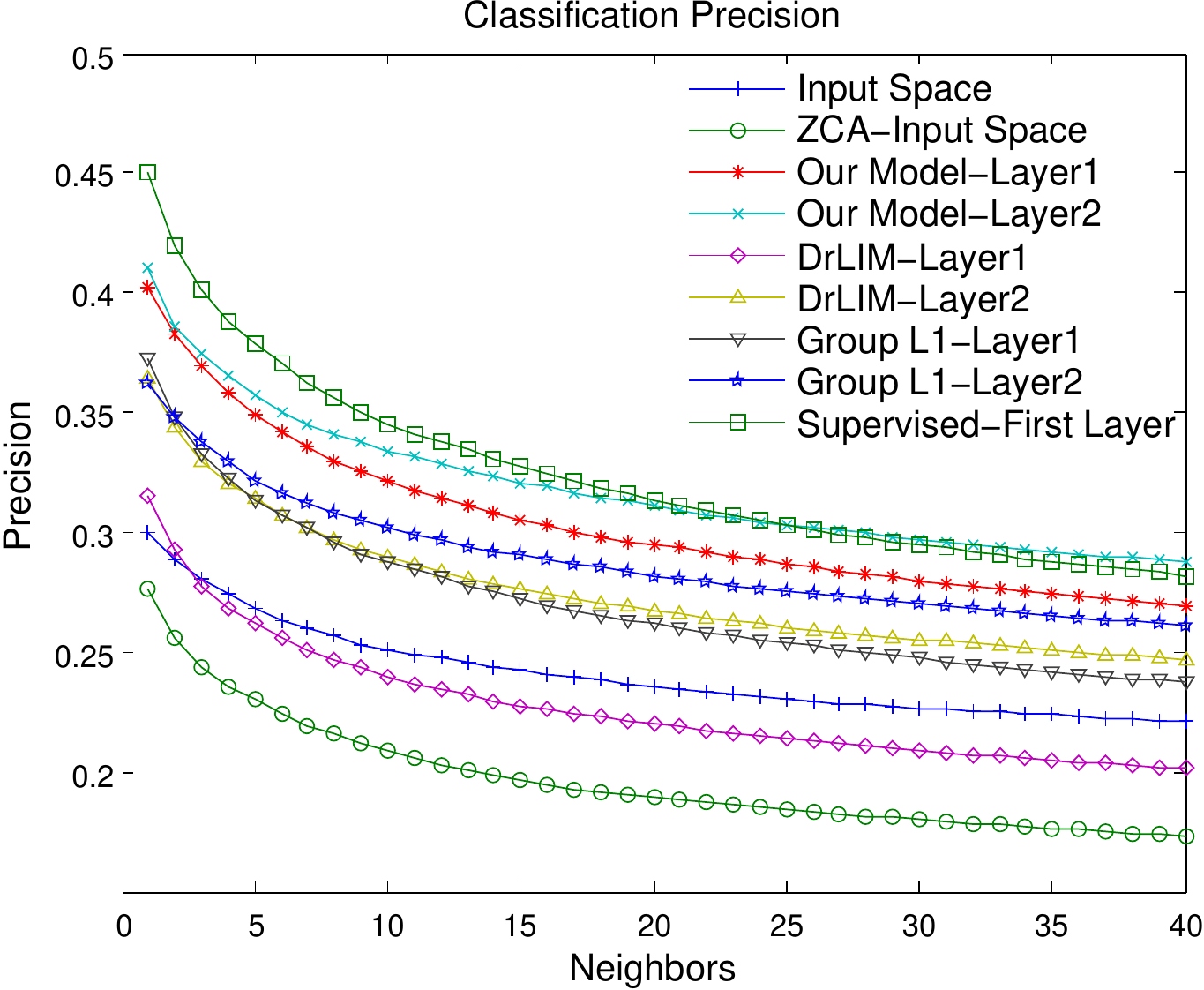}
        \caption{}
        \label{fig:ROCCIFAR}
  \end{subfigure} 
	\caption{Precision-Recall curves corresponding to the YouTube (a) and CIFAR-10 (b) dataset.}
	\label{fig:ROC}
\end{figure}

\begin{figure}
  \centering
\begin{tabular}{| l | l | l | l |} 
\hline 
Model & Optimization & AUC$_{time}$ & AUC$_{class}$ \\ \hline
 Our Model Layer1 & --- & 0.262 & 0.296 \\
 Our Model Layer2 & Greedy & 0.300 & \bf{0.310} \\
 DrLIM Layer1 & --- & 0.188 & 0.221 \\
 DrLIM Layer2 & Joint & \bf{0.378} & 0.268 \\
 Group $L_1$ Layer1 & --- & 0.231 & 0.266 \\
 Group $L_1$ Layer2 & Greedy & 0.285 & 0.281 \\
 \hline
\end{tabular}
\end{figure}

Figure \ref{fig:videoquery} shows the top nine query results on the YouTube dataset for a single frame (left column) in eight spaces. The top row shows the nearest neighbors in pixel space. The second row shows the nearest neighbors in pixel space after ZCA whitening. The next six rows show the nearest neighbors in feature space for one and two layer feature transformations learned with slowness, group sparsity, and DrLIM. The resulting first-layer filters and precision-recall curves are shown in Figures \ref{fig:filters} and \ref{fig:ROC}, respectively.   
Figures \ref{fig:L1filters} and \ref{fig:slowfilters} show the decoders of two one-layer models trained with $\beta = 0,2$, respectively, and a constant value of $\alpha$. The filter bank trained with $\beta = 0$ exhibits no coherence within each pool group; the filters are not visually similar nor do they tend to co-activate at spatially neighboring locations. Most groups in the filter bank trained with slowness tend be visually similar, corresponding to similar colors and/or geometric structures. The features learned by minimizing the DrLIM loss (Equation \ref{eqn:drlimcrit}), which more directly optimizes temporal coherence, have much more high frequency content than the filters learned with any of the auto-encoder methods. Nevertheless, some filters within the same pool group exhibit similar geometric and color structure (Figure~\ref{fig:drlimfilters}). The features learned with a group-sparsity regularizer leads to nearly identical features (and nearly identical activations) within each pool group (Figure~\ref{fig:groupL1filters}). This is not surprising because group sparsity promotes co-activation of the features within each pool group, by definition. We have also tried including an individual sparsity prior, as in Equation~\ref{eqn:loss}, in order to encourage independence among the pooled features. However this has lead to significantly worse temporal-coherence performance.

Figure \ref{fig:cifarquery} shows the result of two queries in the CIFAR-10 dataset. The corresponding precision-recall curves are shown in Figure \ref{fig:ROCCIFAR}. One-layer DrLIM (4096 dimensional) exhibit poor performance in both the temporal and class-based recall tasks. In contrast, jointly trained two-layer DrLIM features (1024 dimensional) exhibit excellent temporal coherence, outperforming all other models by a large margin. Although better than the  first layer, second layer features perform significantly worse on the CIFAR task than even the first-layer features learned by our model. Furthermore, the nearest neighbors in both the one and two-layer feature spaces learned with DrLIM are often neither visually nor semantically similar (see Figure \ref{fig:cifarquery}). The conclusion which can be drawn from this result is that \emph{directly maximizing temporal coherence alone is not a sufficient condition for achieving a semantically (or even visually) coherent features}. However, combining it with reconstruction and sparsity, as in our model, yields the most semantically discriminative features. Although significantly better than the features learned with DrLIM, the features learned with group sparsity exhibit slightly weaker temporal coherence and significantly worse class-based recall. Note that since all the features within a pool group are practically identical, the invariants captured by the pool groups are limited to local translations due to the spatial pooling. As a final comparison, we trained a four layer ConvNet with supervision on CIFAR-10, this network achieved approximately 80\% classification accuracy on the test set. The architecture of the first two stages of the ConvNet is identical to the architecture of the first and second unsupervised stages. The precision curve corresponding to the first layer of the ConvNet is shown in Figure \ref{fig:ROCCIFAR}, which is matched by our-model's second layer at high recall. 

\section{Conclusion}
Video data provides a virtually infinite source of information to learn 
meaningful and complex visual invariances. While temporal slowness is an attractive prior for good visual features, in practice it involves optimizing conflicting objectives that balance invariance and discriminability. In other words, perfectly slow features cannot be informative. An alternative is to replace the small temporal velocity prior with small temporal acceleration, leading to a criteria that \emph{linearizes} observed variability. The resulting representation offers potential advantages, such as extraction of both locally invariant and locally covariant features. Although pooling representations are widespread in visual and audio recognition architectures, much is left to be understood. In particular, a major question is 
how to learn a stacked pooling representation, such that its invariance properties are boosted while controlling the amount of information lost at each layer. This could be possible by replacing the linear decoder of the proposed model with a non-linear decoder which can be used to reconstruct the input from pooled representations. Slow feature learning is merely one way to learn from temporally coherent data. In this work we have provided an auto-encoder formulation of the problem and shown that the resulting features are more stable to naturally occurring temporal variability, while maintaining discriminative power.

{\small
\bibliographystyle{ieee}
\bibliography{ICCV}
}

\end{document}